\pdfoutput=1
\documentclass[11pt]{article}

\usepackage[preprint]{acl}

\usepackage{times}
\usepackage{latexsym}

\usepackage[T1]{fontenc}
\usepackage[utf8]{inputenc}

\usepackage{microtype}

\usepackage{inconsolata}

\usepackage{graphicx}

\usepackage{multirow}
\usepackage{todonotes}
\usepackage{amsmath}
\usepackage{booktabs}
\usepackage{hyperref}
\usepackage{xspace}
\usepackage{subcaption}
\usepackage{listings}
\usepackage{tabularx}
\usepackage{soul}
\newcommand{\mytoprule}{\specialrule{1.15pt}{0em}{0em}}
\newcommand{\NAME}{Big-O\textsubscript{tok}\xspace}
\newcommand{\METRICNAME}{Token Cost\xspace}
\newcommand{\METRICABBREV}{TC\xspace}

\title{Incorporating Token Usage into Prompting Strategy Evaluation}

\author{
 \textbf{Chris Sypherd\textsuperscript{1}},
 \textbf{Sergei Petrov\textsuperscript{2}},
 \textbf{Sonny George\textsuperscript{3}},
 \textbf{Vaishak Belle\textsuperscript{1}}
\\
\\
 \textsuperscript{1}University of Edinburgh,
 \textsuperscript{2}Independent Researcher,
 \textsuperscript{3}Brandeis University
\\
 \small{
   \textbf{Correspondence:} \href{mailto:email@domain}{c.n.sypherd@sms.ed.ac.uk}
 }
}

\begin{document}

\maketitle

%%%%%%%%%%%%%%%%%%%%%%%%%%%%%%%%%%%%%%%%%%%%%%%%%%%%%%%%%%%%%
%%%%%%%%%%%%%%%%%%%%%%%%%%%%%%%%%%%%%%%%%%%%%%%%%%%%%%%%%%%%%

\begin{abstract}
In recent years, large language models have demonstrated remarkable performance across diverse tasks. However, their task effectiveness is heavily dependent on the prompting strategy used to elicit output, which can vary widely in both performance and token usage. While task performance is often used to determine prompting strategy success, we argue that efficiency—balancing performance and token usage—can be a more practical metric for real-world utility. To enable this, we propose \NAME, a theoretical framework for describing the token usage growth of prompting strategies, and analyze \METRICNAME, an empirical measure of tokens per performance. We apply these to several common prompting strategies and find that increased token usage leads to drastically diminishing performance returns. Our results validate the \NAME analyses and reinforce the need for efficiency-aware evaluations.

\end{abstract}

%%%%%%%%%%%%%%%%%%%%%%%%%%%%%%%%%%%%%%%%%%%%%%%%%%%%%%%%%%%%%
%%%%%%%%%%%%%%%%%%%%%%%%%%%%%%%%%%%%%%%%%%%%%%%%%%%%%%%%%%%%%

\section{Introduction} \label{sec:intro}

Large language models (LLMs) are primarily interacted with through natural language prompts. The composition of a prompt exercises significant, often unexpected, influence over the generated output. This has sparked research into "prompt engineering," the study of prompt design to extract maximum performance from LLMs \cite{white2023promptpatterncatalogenhance}. There are many ways to approach prompt engineering; in this paper, we focus on formalized \textbf{prompting strategies}, the overarching paradigms of prompt design (e.g., providing examples of question-answer pairs \cite{FewShotLearners})

As prompting strategies have developed alongside LLMs, benchmark accuracy has emerged as the primary metric for success. New prompting strategies are often tested alongside prior ones on a selection of benchmarks and LLMs, using the gain in accuracy over existing strategies as validation of the new approach. Token usage, if included, is often analyzed post-hoc, indicating that it was only a secondary consideration during development. Optimization for performance without regard for token usage leads to inefficient prompting strategies. Our purpose in this work is to demonstrate another, more holistic approach to prompting strategy evaluation and analysis by (1) proposing \NAME, a framework for comparing theoretical token usage between distinct prompting strategies and (2) introducing \METRICNAME (\METRICABBREV), a simple empirical metric to quantify prompting strategy efficiency.

To achieve these goals, we approach prompting strategy efficiency on two fronts: theoretical and empirical. For our theoretical analysis, we derive \NAME token complexities for a selection of prompting strategies, similar to time complexity analyses common in software engineering. We substantiate our \NAME analyses by evaluating our selection of prompting strategies against common benchmarks using multiple models. We analyze the results of those experiments in terms of \METRICABBREV, to compare how performance and token usage interact. Our results evidence that, while there is performance improvement to be gained from more complex, token-hungry prompting strategies, increasing token usage results in drastically diminishing performance returns.

%%%%%%%%%%%%%%%%%%%%%%%%%%%%%%%%%%%%%%%%%%%%%%%%%%%%%%%%%%%%%
%%%%%%%%%%%%%%%%%%%%%%%%%%%%%%%%%%%%%%%%%%%%%%%%%%%%%%%%%%%%%

\section{Related Work} \label{sec:related}

Although LLM efficiency has been an active area of research for years (see \citet{wan2024efficient}), underwhelming emphasis has been placed on the efficiency of prompting strategies. Techniques have emerged to reduce token usage, such as frameworks that dynamically manage token usage at inference time—e.g., FrugalGPT \cite{chen2023frugalgptuselargelanguage} and LLMLingua \cite{jiang2023llmlinguacompressingpromptsaccelerated}—and prompt compression—e.g., \citet{mu2023learninggist}. These approaches seek to enable efficient LLM usage in spite of inefficient prompting strategies, whereas this work promotes a focus on prompting strategy efficiency.

Some work, such as \citet{sivarajkumar2023empiricalevaluationpromptingstrategies, kim-etal-2023-better, chu-etal-2024-navigate}, has been done to evaluate prompting strategies in specific domains, but token usage statistics are often not included. Prompting strategy evaluation is largely left to new prompting strategy proposals (e.g., CoT \cite{wei2022chain}) or benchmarks (e.g., BBH \cite{suzgun2022challengingbbh}). These evaluations tend to prioritize benchmark accuracy without significant consideration of token usage. In this work, we demonstrate the relevance of an increased focus on token efficiency and how to proactively incorporate it into prompting strategy analysis.

Some recent prompting strategies, such as Constrained-CoT \cite{nayab2024concisethoughtsimpactoutput}, Concise-CoT \cite{renze2024benefitsconcisechainthought}, and Algorithm-of-Thoughts \cite{algorithm-thought-sel}, are designed as optimizations over extant strategies with token usage reduction as a primary motivation. Our hope for this work is that it will enable future prompting strategy development and analysis that similarly prioritizes efficiency.

%%%%%%%%%%%%%%%%%%%%%%%%%%%%%%%%%%%%%%%%%%%%%%%%%%%%%%%%%%%%%
%%%%%%%%%%%%%%%%%%%%%%%%%%%%%%%%%%%%%%%%%%%%%%%%%%%%%%%%%%%%%

\section{Methodology}

We explore the importance of token usage both theoretically and empirically. Due to the popularity of LLMs, there exists an infeasible number of possible evaluation combinations\footnote{Prompting strategies: 40+ \cite{vatsal2024surveypromptengineeringmethods, chu-etal-2024-navigate}; Benchmarks: 130+ \cite{eval-harness}; Open-source, benchmarked LLMs: 3200+, as of January, 2025 \cite{open-llm-leaderboard-v2}.}. To focus the scope of this paper on token usage, we restrict the number of prompting strategies, benchmarks, and models we use. We discuss our selection processes in Sections \ref{sec:theor-anal} and \ref{sec:prac-eval}.

%%%%%%%%%%%%%%%%%%%%%%%%%%%%%%%%%%%%%%%%%%%%%%%%%%%%%%%%%%%%%

\subsection{Theoretical Analysis} \label{sec:theor-anal}

\begin{table}[h]
  \centering
  \small
  \begin{tabular}{p{57pt}|p{25pt}|p{100pt}}
    \hline
    \textbf{Prompting Strategy} & \textbf{\NAME} & \textbf{Variables} \\\mytoprule
    Vanilla IO & $O(1)$ & \\
    Zeroshot CoT \newline \tiny \cite{kojima2022largezeroshotcot} & $O(1)$ & \\
    Vanilla Fewshot \newline \tiny \cite{FewShotLearners} & $O(k)$ & $k$: k-shot exemplars \\
    Fewshot CoT \newline \tiny \cite{wei2022chain} & $O(k)$ & $k$: k-shot exemplars \\
    CoT-SC \newline \tiny \cite{wang2023selfconsistency} & $O(pk)$& $k$: k-shot exemplars; $p$: sampled chains \\\mytoprule
  \end{tabular}
  \caption{\NAME token complexities for each prompting strategy.}
  \label{tab:complex}
\end{table}

We categorize prompting strategies into three broad groups: \textbf{(1) linguistic prompt engineering}, which relies on specific phrasing techniques—e.g., Plan-and-Solve \cite{wang-etal-2023-plan} or Zeroshot CoT \cite{kojima2022largezeroshotcot}; \textbf{(2) in-context learning}, which consists of providing examples of task-response pairs before providing the task to the LLM—e.g., Vanilla Fewshot \cite{FewShotLearners} or Fewshot CoT \cite{wei2022chain}; and \textbf{(3) multi-hop}, which is characterized by multiple LLM calls—e.g., Least-To-Most \cite{zhou2023leasttomost}, Tree-of-Thought \cite{yao2023tree}, or CoT Self-Consistency\footnote{Abbreviated as CoT-SC\textsubscript{$n$}, where $n$ is the number of sampled chains.} \cite{wang2023selfconsistency}. These 3 prompting strategy categories roughly correspond to the following 3 \NAME complexity classes, respectively: \textbf{(1)} constant—e.g., the consistent overhead of "Think step by step" \cite{wei2022chain}; \textbf{(2)} linear—e.g., the number of fewshot exemplars; and \textbf{(3)} polynomial or higher—e.g., the number of multi-hop steps times the number of exemplars. These \NAME complexity classes are reflected in Table \ref{tab:complex}.

To ensure our investigation represents all 3 categorizations, we select prompting strategies from each. Namely, we choose Vanilla IO (i.e., simply providing the benchmark question) as a baseline; Zeroshot CoT to represent \textbf{(1)}; Vanilla Fewshot and Fewshot CoT for \textbf{(2)}; and CoT-SC for \textbf{(3)}. These strategies are widely adopted, tend to build on each other without significant changes to prompt design, and demonstrate an organic evolution of prompting strategies over several years \cite{chu-etal-2024-navigate}.

The purpose of \NAME is to provide an objective representation of the theoretical token usage of a given prompting strategy, enabling direct comparison with the \NAME of other prompting strategies. \NAME is based on Big-O notation \cite{big-o1976} and thus we rely on the terminology and definitions associated with it.

\NAME describes the asymptotic growth of token usage as a function of variables\footnote{We treat the initial input that the prompting strategy modifies (e.g., a benchmark question) to be constant and exclude it for simplicity. Similarly, we treat additive adjustments to the input or output (e.g., CoT's "Think step by step" \cite{wei2022chain}) as constants.} in the prompting strategy (e.g., the number of fewshot examples). It is derived analogously to Big-O time complexity: by considering how token usage increases as prompting strategy variables approach infinity. The variables with the highest growth rate dominate the other terms (e.g., constants, lower-order variables, and scalars), which can then be omitted. \NAME token complexity can often be derived from a natural language description of a prompting strategy. We provide sample derivations for the \NAME functions from Table \ref{tab:complex} in Appendix \ref{app:theor-deriv}.

%%%%%%%%%%%%%%%%%%%%%%%%%%%%%%%%%%%%%%%%%%%%%%%%%%%%%%%%%%%%%

\subsection{Empirical Analysis} \label{sec:prac-eval}

We test our selection of prompting strategies against 3 common benchmarks using 3 LLMs. To perform the empirical evaluations, we leverage LM Evaluation Harness \cite{biderman2024lessonstrenchesreproducibleevaluation, eval-harness}, a framework aimed at increasing the reproducibility of LLM evaluations.

We base our selection of models on recency, popularity, and size. We do not use commercial models due to budget constraints\footnote{See Appendix \ref{app:commercial-cost} for cost estimates for commercial APIs.}. We select Llama 3.1 8B Instruct \cite{dubey2024llama3herdmodels}, Qwen 2.5 14B Instruct, and Qwen 2.5 32B Instruct \cite{qwen2025qwen25technicalreport}. This selection provides coverage of various sizes of smaller models (each approximately doubling the parameter count of the prior) and diversity of origin, to ensure multiple approaches to data collection, training, and alignment are represented\footnote{We choose two Qwen 2.5 models to facilitate a comparison between model size, found in Appendix \ref{app:abl:size}.}.

For benchmarks, we select BBH \cite{suzgun2022challengingbbh}, GSM8K \cite{gsm8k}, and MMLU \cite{hendryckstest2021mmlu}. This represents a diverse group of general-purpose benchmarks based on typical accuracy ranges\footnote{GSM8K: ~80-95\%; BBH: ~50-87\%; MMLU: ~70-92\% \cite{open-llm-leaderboard-v2, dubey2024llama3herdmodels, qwen2025qwen25technicalreport}.} and response type\footnote{GSM8K: free response number; BBH: free response text; MMLU: multiple choice.}. For fewshot prompting strategies, we use 3 exemplars for BBH, 8 for GSM8K, and 4 for MMLU\footnote{These numbers are based on the availability of CoT examples in LM Evaluation Harness and closely reflect the number of examples suggested in CoT-SC \cite{wang2023selfconsistency}.}.

%%%%%%%%%%%%%%%%%%%%%%%%%%%%%%%%%%%%%%%%%%%%%%%%%%%%%%%%%%%%%
%%%%%%%%%%%%%%%%%%%%%%%%%%%%%%%%%%%%%%%%%%%%%%%%%%%%%%%%%%%%%

\section{Results} \label{sec:results}

\begin{table}[ht]
  \centering
  \small
  \setlength{\tabcolsep}{3.25pt}
  \begin{tabular}{p{58pt}|c|c|c|c|c}
    \hline
    \textbf{Column:} Higher Token Usage (Numerator)\newline\textbf{Row:} Lower Token Usage (Denominator) & \rotatebox[origin=r]{90}{CoT-SC\textsubscript{10}} & \rotatebox[origin=r]{90}{CoT-SC\textsubscript{5}} & \rotatebox[origin=r]{90}{Fewshot CoT} & \rotatebox[origin=r]{90}{Vanilla Fewshot} & \rotatebox[origin=r]{90}{Zeroshot CoT}\\\mytoprule
    Vanilla IO & \textcolor{blue}{50}; \textcolor{orange}{29.3} & \textcolor{blue}{25}; \textcolor{orange}{14.6} & \textcolor{blue}{5}; \textcolor{orange}{3.0} & \textcolor{blue}{5}; \textcolor{orange}{2.2} & \textcolor{blue}{1}; \textcolor{orange}{1.3} \\
    Zeroshot CoT & \textcolor{blue}{50}; \textcolor{orange}{23.4} & \textcolor{blue}{25}; \textcolor{orange}{11.7} & \textcolor{blue}{5}; \textcolor{orange}{2.4} & \textcolor{blue}{5}; \textcolor{orange}{1.7} & \\
    Vanilla Fewshot & \textcolor{blue}{10}; \textcolor{orange}{13.5} & \textcolor{blue}{5}; \textcolor{orange}{6.8} & \textcolor{blue}{1}; \textcolor{orange}{1.4} & & \\
    Fewshot CoT & \textcolor{blue}{10}; \textcolor{orange}{9.9} & \textcolor{blue}{5}; \textcolor{orange}{5.0} &  & & \\
    CoT-SC\textsubscript{5} & \textcolor{blue}{2}; \textcolor{orange}{2.0} &  &  & & \\\mytoprule
  \end{tabular}
  \caption{Theoretical and observed token usage ratios between prompting strategies, averaged over the 3 benchmarks. Values are formatted as \textcolor{blue}{\textit{<theoretical>}}\textit{; }\textcolor{orange}{\textit{<observed>}} and derived by $\frac{num\ tokens_2}{num\ tokens_1}$, where $num\ tokens_2 > num\ tokens_1$.}
  \label{tab:multiple-comp}
\end{table}

\begin{figure*}[h]
    \centering
    \begin{minipage}[b]{0.32\textwidth}
        \centering
        \includegraphics[width=\textwidth]{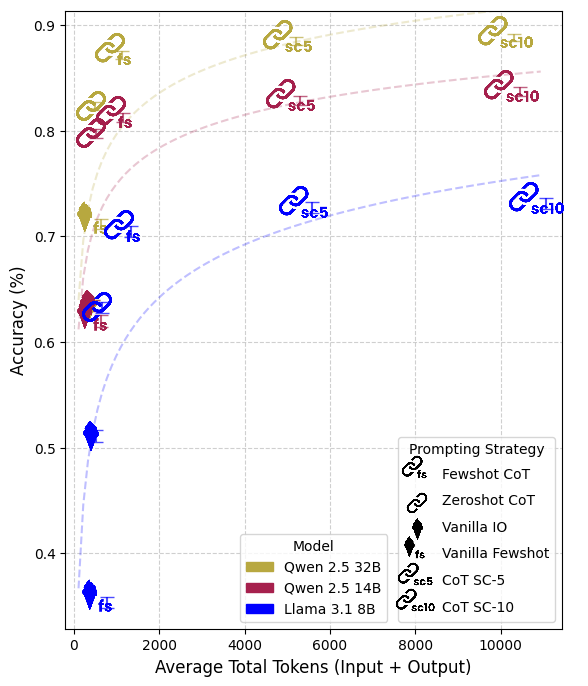}
        \small{(A) BBH}
    \end{minipage}
    \begin{minipage}[b]{0.32\textwidth}
        \centering
        \includegraphics[width=\textwidth]{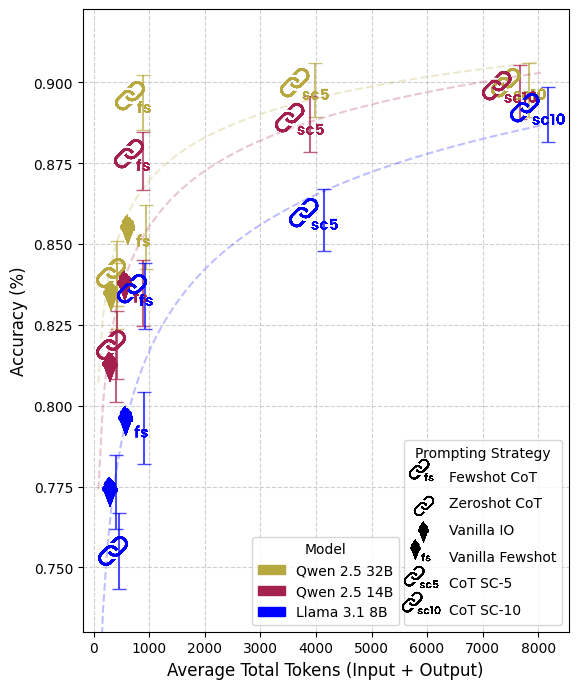}
        \small{(B) GSM8K}
    \end{minipage}
    \begin{minipage}[b]{0.32\textwidth}
        \centering
        \includegraphics[width=\textwidth]{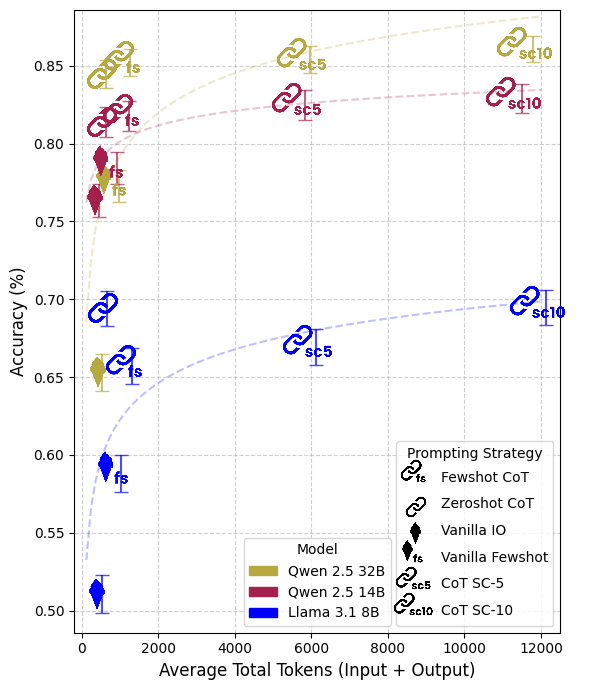}
        \small{(C) MMLU}
    \end{minipage}
    \caption{Accuracy vs. token usage plots with standard error bars for various prompting strategies, models, and benchmarks. The trend lines demonstrate the rapid growth of \METRICABBREV for these strategies.}
    \label{fig:three_plots}
\end{figure*}

To substantiate our \NAME analyses, we use the observed token usages from our experiments to calculate the relative token usage ratios between prompting strategies. We derive theoretical estimates of those ratios from our \NAME functions by substituting in the values from our experiments for the variables in \NAME (e.g., $p=5$ and $\bar{k}=5$ for CoT-SC\textsubscript{5}). The results of that comparison are found in Table \ref{tab:multiple-comp}. We expect noise in the observed token usage due to: inherent token usage (e.g., chat templates); the relatively low values of prompting strategy variables (e.g., $k=3$ fewshot exemplars for BBH); and the unpredictability of LLM output. However, while the observed and theoretical factors are not perfect matches, our findings do correctly align with the \NAME token complexity classes discussed in Section \ref{sec:theor-anal} (constant, linear, and polynomial). In other words,
\begin{displaymath}
T\textsubscript{O\textsubscript{tok}(1),b,m}()<T\textsubscript{O\textsubscript{tok}(k),b,m}(k)<T\textsubscript{O\textsubscript{tok}(pk),b,m}(p,k)
\end{displaymath}
where $T\textsubscript{O\textsubscript{tok},b,m}(x)$ is the observed token usage for each prompting strategy in the O\textsubscript{tok} complexity class, benchmark $b$, and model $m$ in our experiments.

To quantify token efficiency, we discuss the results from our experiments in terms of \METRICNAME\footnote{See Appendix \ref{app:interpret-tc} for an expanded discussion on \METRICABBREV interpretation, including edge cases.} (\METRICABBREV). We define \METRICABBREV as the number of tokens\footnote{Token counts are estimated by $\frac{\text{num\ characters}}{4}$.} per percentage point of accuracy (expressed as $\frac{t}{p}$). The inverse of \METRICABBREV can be thought of as token efficiency; thus, relatively high \METRICABBREV is less efficient while lower \METRICABBREV is more efficient. We use this metric, $\frac{t}{p}$, rather than the inverse, $\frac{p}{t}$, because we find it to be more intuitive in the context of prompting strategies.

The results of the experiments outlined in Section \ref{sec:prac-eval} are found in Figure \ref{fig:three_plots}. Across all benchmarks and models, our experiments demonstrate consistent trend lines (of the form $y=\log(\log(x))$), reflecting the diminishing accuracy returns from increased token usage. In other words, it requires significantly more tokens to realize performance gains as token usage increases. To discuss this trend in terms of \METRICABBREV, we explore both \textit{average} and \textit{marginal} \METRICABBREV\footnote{We use the average of ratios when discussing each to give equal weighting to every observation.}. Average \METRICABBREV is simply the token usage divided by the accuracy for a given prompting strategy ($\frac{num\ tokens_{obs}}{accuracy_{obs}}$). Marginal \METRICABBREV is the change in token usage to realize the change in performance between two prompting strategies. In other words, for $num\ tokens_2 >= num\ tokens_1$,
\[\frac{num\ tokens_2 - num\ tokens_1}{accuracy_2 - accuracy_1}\]
Both average and marginal \METRICABBREV can be thought of as the slope between two points (one being the origin, for average \METRICABBREV), which represents the expected cost, in tokens, of adding one point of accuracy.

Across all experiments, the average \METRICABBREV for the prompting strategy with the lowest accuracy is 5.0 $\frac{t}{p}$, while that of the highest performing prompting strategy is 119.4 $\frac{t}{p}$, a more than 20x increase in average \METRICABBREV and, inversely, 20x \textit{decrease} in efficiency. This is reflected in the plots in Figure \ref{fig:three_plots} in that even the worst performing prompting strategies still attain relatively high accuracy and more complex ones make only small gains over that for vastly more token usage. Using accuracy as the sole metric ignores the drastic decrease in efficiency that can result from increased token usage.

In Figure \ref{fig:three_plots}, we observe an initially steep curve, indicating that tokens are traded relatively efficiently for accuracy, followed by a plateau, where tokens are traded more inefficiently for accuracy. To compare the marginal \METRICABBREV along this curve, we select Vanilla IO, Fewshot CoT, and CoT-SC\textsubscript{10} which tend to lie on the extremes of the trend lines. Across all experiments, the marginal \METRICABBREV between Vanilla IO and Fewshot CoT is 65.3 $\frac{t}{p}$ while the marginal \METRICABBREV between Fewshot CoT and CoT-SC is 6701.8 $\frac{t}{p}$, a decrease in efficiency of more than two orders of magnitude. In a high-stakes scenario where accuracy is paramount, pursuing performance gains at a rate of 6701.8 $\frac{t}{p}$ may be an acceptable cost. For many real-world scenarios, increasing accuracy at a rate of 65.3 $\frac{t}{p}$ might be more reasonable. \METRICABBREV provides an intuitive way to compare the tradeoff between token usage and performance and allows for more informed prompting strategy and variable selection.

We perform an ablation study, found in Appendix \ref{app:abl:fewshot}, on the number of fewshot exemplars. We find that incrementally increasing the number of fewshot examples follows a similar trend of yielding diminishing returns as token usage increases.

All information for reproducing our results, as well as our verbatim results, are detailed in Appendices \ref{app:repro} and \ref{app:configs}.

\section{Conclusion}

Token usage represents a significant, yet often underrepresented, component of prompting strategy evaluation. To facilitate the comparison of token usage between distinct prompting strategies, we present \NAME token complexity and substantiate it empirically by comparing predicted token usages to those derived from our experiments. We analyze our experiments in terms of \METRICNAME and use it to demonstrate the tradeoffs between token usage and performance. Our experiments demonstrate the importance of including token usage in prompting strategy evaluation and validate \NAME and \METRICNAME as viable means of doing so.

%%%%%%%%%%%%%%%%%%%%%%%%%%%%%%%%%%%%%%%%%%%%%%%%%%%%%%%%%%%%%
%%%%%%%%%%%%%%%%%%%%%%%%%%%%%%%%%%%%%%%%%%%%%%%%%%%%%%%%%%%%%

\section*{Limitations} \label{limits}

To focus the contribution of this work, we make a number of thoughtful concessions, such as the models, benchmarks, and prompting strategies we use. We explicitly justify the most relevant limitations in the main text above--such as in Sections \ref{sec:theor-anal} and \ref{sec:prac-eval}--and note other minor limitations here to further demonstrate the purpose and scope of this work. Despite the limitations, we maintain that the core premise of this work--demonstrating how to incorporate token usage into prompting strategy evaluation--remains broadly applicable beyond our experiments.

\paragraph{Prompting Strategies.}

In Section \ref{sec:intro}, we narrow the scope of this work to a subset of the broader prompt engineering landscape: formalized prompting strategies. As detailed throughout the Appendix, we strive to control for factors extraneous to the minimalist instantiation of each prompting strategy. We do so to isolate the effects of token usage dictated by the prompting strategies and the resultant benchmark performance to demonstrate how a significant area of research has become prone to inefficiencies due to a lack of relevant metrics. Although we focus on formalized prompting strategies, \NAME and \METRICABBREV are useful metrics for other aspects of prompt engineering as well, such as linguistic and language choices.

As noted in Sections \ref{sec:intro} and \ref{sec:related}, the focus of this work is on incorporating token usage into prompting strategy evaluationand analysis. We do not seek to seek to solve issues of prompting strategy efficiency but instead provide methods for quantifying it, both for researchers and practitioners. To maintain that scope, we do not explore nor propose specific methods of optimizing prompts and instead focus on introducing insightful metrics and demonstrating their utility in practice.

The results we present here represent a single thread of prompting strategy evolution; we expect the results to follow a predictable pattern (e.g., diminishing accuracy returns) because there are no drastic changes to the principles underlying our selection of prompting strategies. It is likely that fundamentally different prompting strategies, such as Least-to-Most \cite{zhou2023leasttomost} or Algorithm of Thoughts \cite{algorithm-thought-sel}, would not follow the trend lines in our plots.

\paragraph{Models.}

Depending on the data collection and processing methods used by LLM creators, benchmark data leakage could influence our empirical results, as demonstrated by \citet{gsm-symbolic}. LLMs that were trained on data that included BBH question-answer pairs, for example, could be influenced by prompting strategies to a lesser degree than those that were not. We use models from multiple sources in conjunction with multiple benchmarks to mitigate the potential effects of data leakage.

In this work, we exclusively consider autoregressive, text-to-text ("traditional") LLMs because most prompting strategies are optimized for them. We recognize, however, that multimodal and, more recently, reasoning LLMs have become increasingly relevant \cite{deepseekai,multimodal-survey-yin,caffagni-multimodal}. We exclude them from our investigation here for a number of reasons: (1) prompt engineering specific to such models is a nascent field and distinct from prompt engineering for traditional LLMs\footnote{For reasoning models, some commercial providers even advise against the use of established prompting strategies (see \url{https://platform.openai.com/docs/guides/reasoning-best-practices}).} \citep{wu2024visualpromptingmultimodallarge}; (2) due to the recency of reasoning models, there are very few (especially open-source) models available; and (3) the inclusion of multimodal benchmarks and the use of reasoning models would drastically increase the compute required to undertake a similar study. Nevertheless, we maintain that the efficiency-aware metrics explored here remain relevant to such models since they function simply on tokenized inputs and outputs. We see prompting strategies designed for multimodal and, particularly, reasoning LLMs as a significant avenue for future research and are hopeful that \NAME and \METRICABBREV will be incorporated into their development.

While we believe it a fair comparison that lends itself to real-world deployment, we recognize that running our selection of prompting strategies with relatively small LLMs on a subset of benchmarks does not fully reflect the performance of the strategy under all conditions. It is very likely, for example, that the CoT prompting strategy \cite{wei2022chain} would be leveraged better by Llama 3.1 405B than by Llama 3.1 8B, due to the former exhibiting superior reasoning capabilities \cite{dubey2024llama3herdmodels}. The strength of an LLM may magnify the disparity between a specific prompting strategy and others.

\paragraph{Benchmarks.}

Similarly, while we attempt to cover a breadth of domains in our selection of benchmarks, this selection may fail to highlight the strengths of some prompting strategies over others in certain domains. Our purpose is not to rank prompting strategies but to demonstrate the tradeoff between token usage and benchmark accuracy. We recognize, however, that our selection of benchmarks may fail to cover the strength of a particular prompting strategy entirely, which may paint it in a worse light than it deserves. To mitigate this point, we focus on generalist prompting strategies and benchmarks and detail our selection processes in Sections \ref{sec:theor-anal} and \ref{sec:prac-eval}.

A common issue in LLM benchmarking is reliable answer extraction. Often, regular expressions are used, which are not robust to the unpredictable output formats an LLM may generate. We rely largely on the extraction methods from LM Evaluation Harness but observe certain inaccuracies in answer extraction. This is an open problem in LLM research \cite{anonymous2025xfinder}. For this project, we rely on consistency in answer extraction methods between experiments but recognize that certain correct answers may be marked incorrectly.

\paragraph{\NAME.}

A minor limitation of \NAME is a lack of differentiation between input and output tokens. It is inherently more expensive for an LLM to generate output tokens than to process input tokens due to its autoregressive nature, a fact that is reflected in the pricing structure of common commercial APIs\footnote{E.g., \url{https://openai.com/api/pricing/}}. We considered that differentiating between input and output tokens would have introduced excessive complexity to \NAME, particularly since it does not serve as a precise measure. We instead consider combined token usage (input \textbf{and} output) to provide a holistic view of token consumption.

Another limitation of \NAME is the decreased range of prompting strategy variable values. While variables in traditional Big-O analyses often span many orders of magnitude, variables in prompting strategies tend to be lower (typically $<=100$) \cite{FewShotLearners,wang2023selfconsistency}. As noted in Section \ref{sec:results}, the relatively low values for those variables could result in extraneous factors (e.g., chat templates, model idiosyncrasies, etc.) limiting their impact on overall token usage. However, we observed that, even for extremely low values (e.g., $k=3$ fewshot examples for BBH), the token usages observed in each experiment aligned with the expected \NAME token complexity classes. Although not a precise measure, \NAME can still provide useful insights for expected prompting strategy token usage, even for small values.

%%%%%%%%%%%%%%%%%%%%%%%%%%%%%%%%%%%%%%%%%%%%%%%%%%%%%%%%%%%%%
%%%%%%%%%%%%%%%%%%%%%%%%%%%%%%%%%%%%%%%%%%%%%%%%%%%%%%%%%%%%%

\section*{Ethical Impact}

LLM usage incurs real-world monetary and environmental costs \cite{green-ai,dhar2020carbon,wu2022sustainable}. This work promotes the consideration of token usage in prompting strategy development and evaluation to increase the long-term efficiency of LLM inference.

%%%%%%%%%%%%%%%%%%%%%%%%%%%%%%%%%%%%%%%%%%%%%%%%%%%%%%%%%%%%%
%%%%%%%%%%%%%%%%%%%%%%%%%%%%%%%%%%%%%%%%%%%%%%%%%%%%%%%%%%%%%

\section*{Acknowledgments}

This work was partly funded by a Royal Society University Research Fellowship and a grant from Cisco.

%%%%%%%%%%%%%%%%%%%%%%%%%%%%%%%%%%%%%%%%%%%%%%%%%%%%%%%%%%%%%
%%%%%%%%%%%%%%%%%%%%%%%%%%%%%%%%%%%%%%%%%%%%%%%%%%%%%%%%%%%%%

\bibliography{comp-eff}

%%%%%%%%%%%%%%%%%%%%%%%%%%%%%%%%%%%%%%%%%%%%%%%%%%%%%%%%%%%%%
%%%%%%%%%%%%%%%%%%%%%%%%%%%%%%%%%%%%%%%%%%%%%%%%%%%%%%%%%%%%%

\appendix

\section{\NAME Analysis} \label{app:theor-deriv}

We provide an expanded token complexity analysis for each prompting strategy examined in the main text in Table \ref{tab:full-complex}. We define the minimally viable IO pair (MVIO) for a given benchmark question to be the full text of the question and the minimum amount of text to convey the answer (e.g., Question: "How many days are there in a week?"; Answer: "7"). All other prompting strategies that incur additive adjustments to the input or output (e.g., the natural language thinking induced by CoT's "Think step by step" \cite{wei2022chain}) are treated as constant overheads on top of the MVIO, which are represented by Greek letters.

Table \ref{tab:theor-complex} shows theoretical token usage ratios based on \NAME. The expected token usage ratios used in Table \ref{tab:multiple-comp} are based on these, averaged across the 3 benchmarks.

\begin{table*}
  \centering
  \small
  \begin{tabular}{p{58pt}|p{33pt}|p{75pt}|p{128pt}|p{60pt}}
    \hline
    \textbf{Prompting Strategy} & \textbf{\NAME} & \textbf{Token Complexity} & \textbf{Variables} & \textbf{Values Used in Original Paper} \\\mytoprule
    MVIO    & $O(1)$ & 1 &  &            \\
    Vanilla IO    & $O(1)$ & $1 + \psi$ &  &            \\
    Zeroshot CoT \newline \tiny \cite{kojima2022largezeroshotcot} & $O(1)$ & $1 + \alpha$ &  & \\
    Vanilla Fewshot \newline \tiny \cite{FewShotLearners} & $O(k)$ & $1 + k$ & $k$: k-shot exemplars & k=0, 1, 10-100           \\
    Fewshot CoT \newline \tiny \cite{wei2022chain} & $O(k)$ & $1 + \alpha + k + k\alpha$ & $k$: k-shot exemplars & k=8 \\
    CoT-SC \newline \tiny \cite{wang2023selfconsistency}     & $O(pk)$ & $p(1 + \alpha + k + k\alpha)$ & $k$: k-shot exemplars; $p$: sampled chains & k=4-8; p=40 \\
    % ToT (BFS, Value, Propose) \newline \tiny \cite{yao2023tree} & $O(b^t(k_p + d_ps_vk_v))$ &  &  & b=5; t=3; $d_p$=3; $s_v$=3; $k_p$=1; $k_v$=8? \\
    \mytoprule
  \end{tabular}
  \caption{The theoretical token complexities for various prompting strategies. Greek letters represent the overhead associated with an IO pair (assumed constant per prompting strategy) and Roman letters represent variables.}
  \label{tab:full-complex}
\end{table*}

\begin{table}
  \centering
  \small
  \begin{tabular}{c|c|c|c}
    \hline
    \multirow{2}{*}{\textbf{Prompting Strategy}} & \multicolumn{3}{c}{\textbf{Token Usage Growth Rate}} \\\cline{2-4}
    & \textbf{BBH} & \textbf{GSM8K} & \textbf{MMLU}  \\\mytoprule
    Vanilla IO & 1 & 1 & 1 \\
    CoT Zeroshot & 1 & 1 & 1 \\
    Vanilla Fewshot & 3 & 8 & 4 \\
    Fewshot CoT & 3 & 8 & 4 \\
    CoT-SC\textsubscript{5} & 15 & 40 & 20 \\
    CoT-SC\textsubscript{10} & 30 & 80 & 40 \\\mytoprule
  \end{tabular}
  \caption{The token usage growth rate over MVIO per prompting strategy, derived from Table \ref{tab:complex} using the variables used in our experiments.}
  \label{tab:theor-complex}
\end{table}

\begin{figure*}
\centering
    \subfloat[MVIO]{\label{fig:theor-deriv-a}\includegraphics[width=60mm]{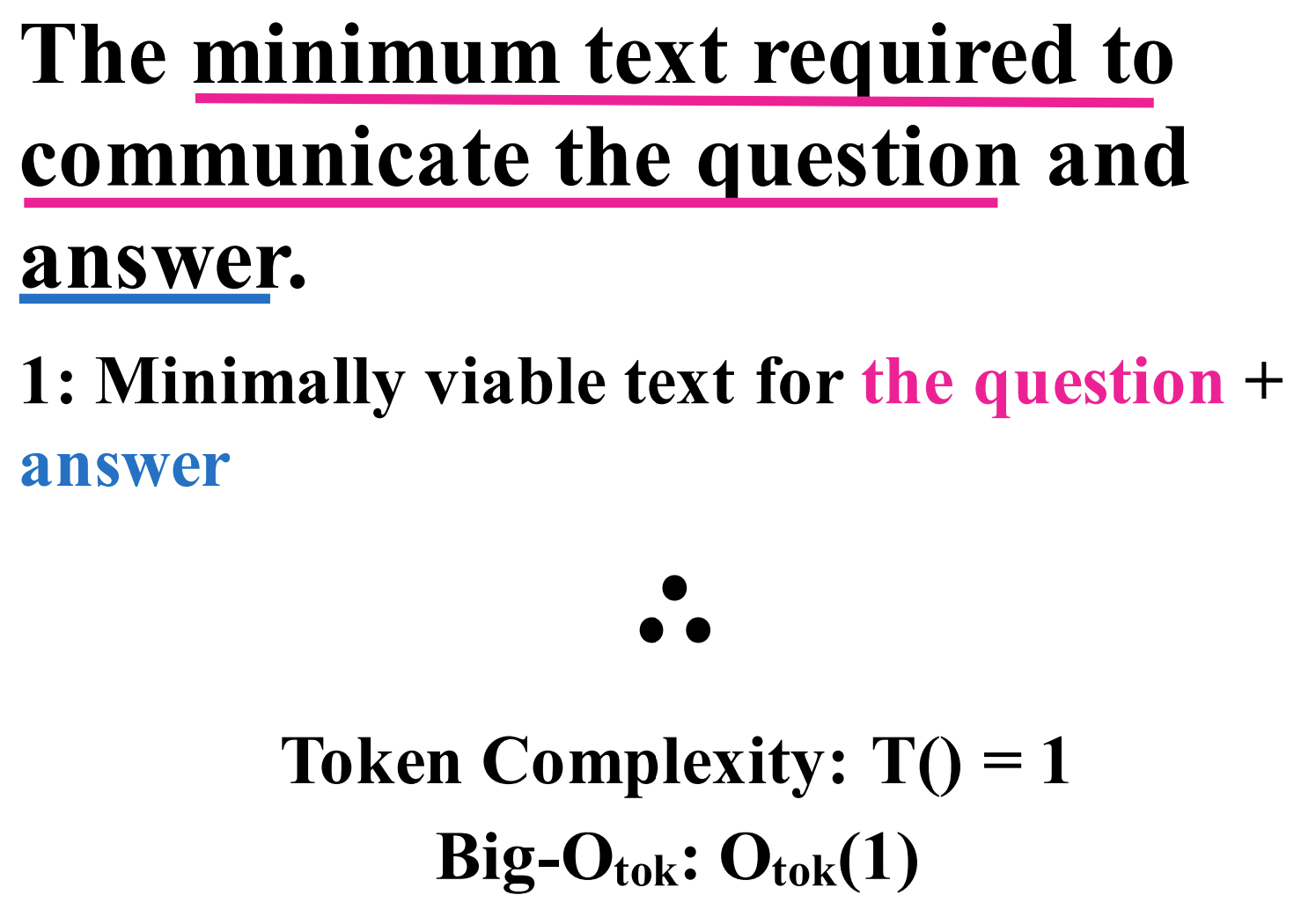}}
    \hspace{1cm}
    \subfloat[Vanilla IO]{\label{fig:theor-deriv-b}\includegraphics[width=60mm]{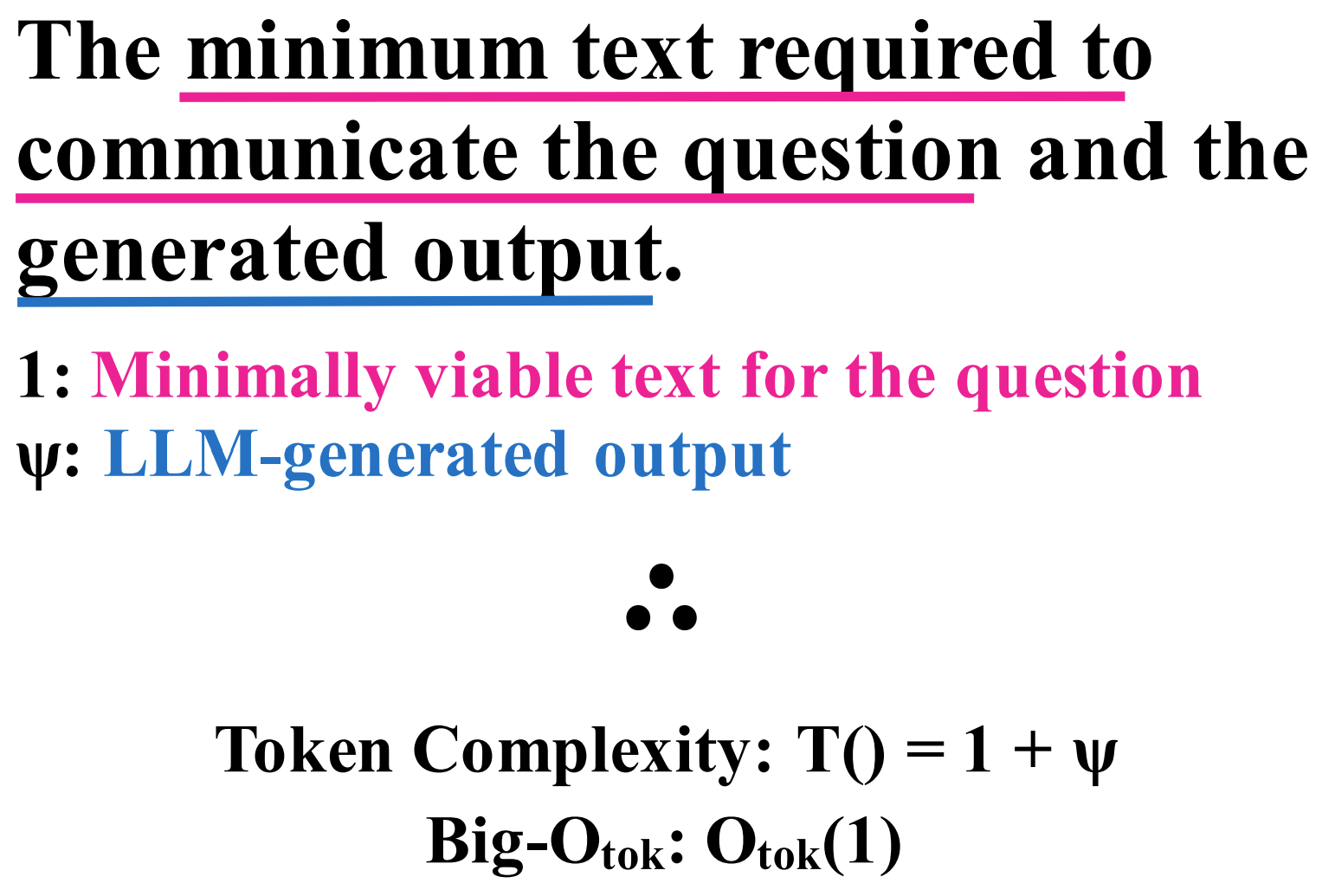}}
    \hspace{1cm}
    \subfloat[Zeroshot CoT]{\label{fig:theor-deriv-c}\includegraphics[width=60mm]{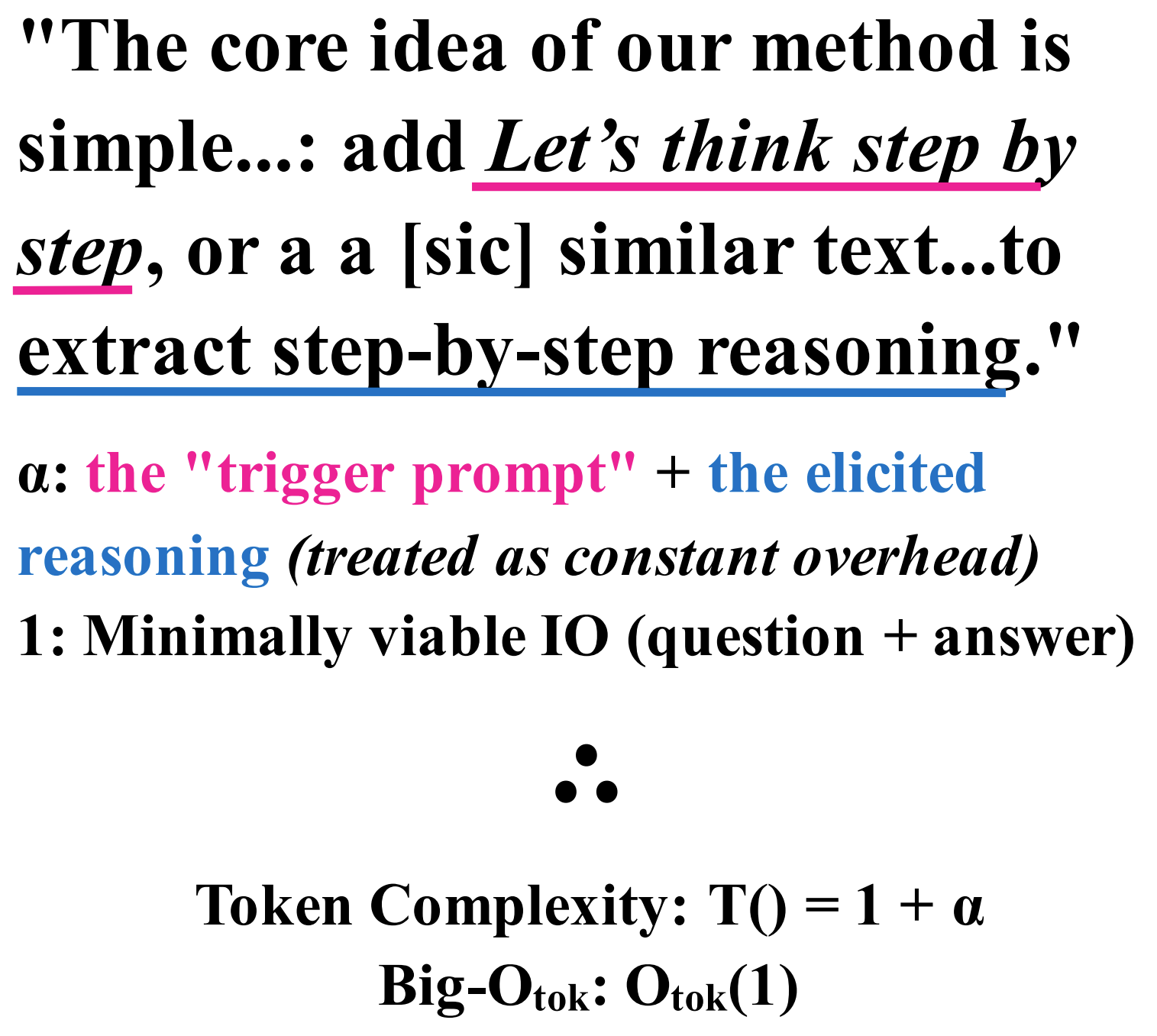}}
    \hspace{1cm}
    \subfloat[Vanilla Fewshot]{\label{fig:theor-deriv-d}\includegraphics[width=60mm]{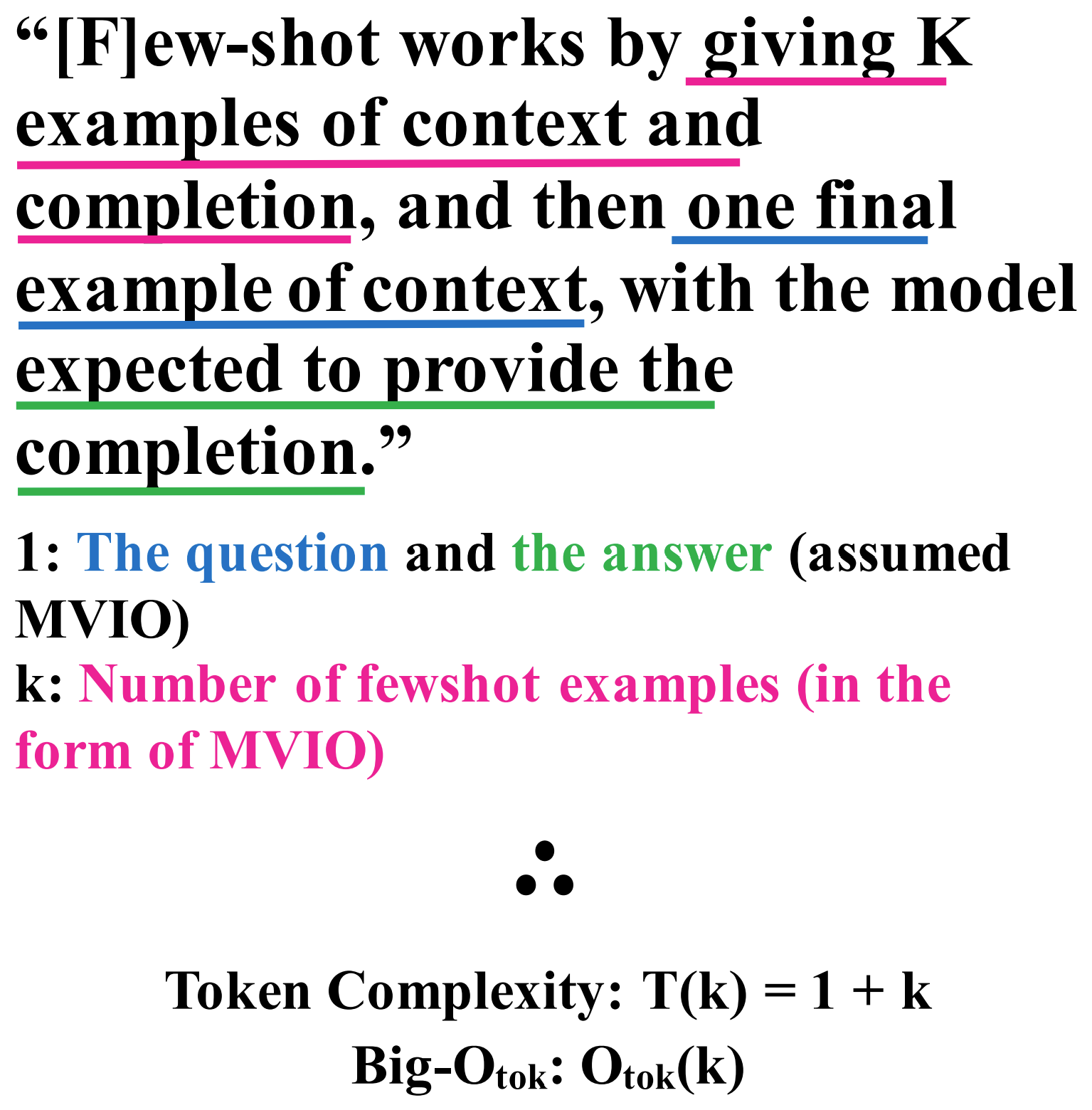}}
    \hspace{1cm}
    \subfloat[Fewshot CoT]{\label{fig:theor-deriv-e}\includegraphics[width=60mm]{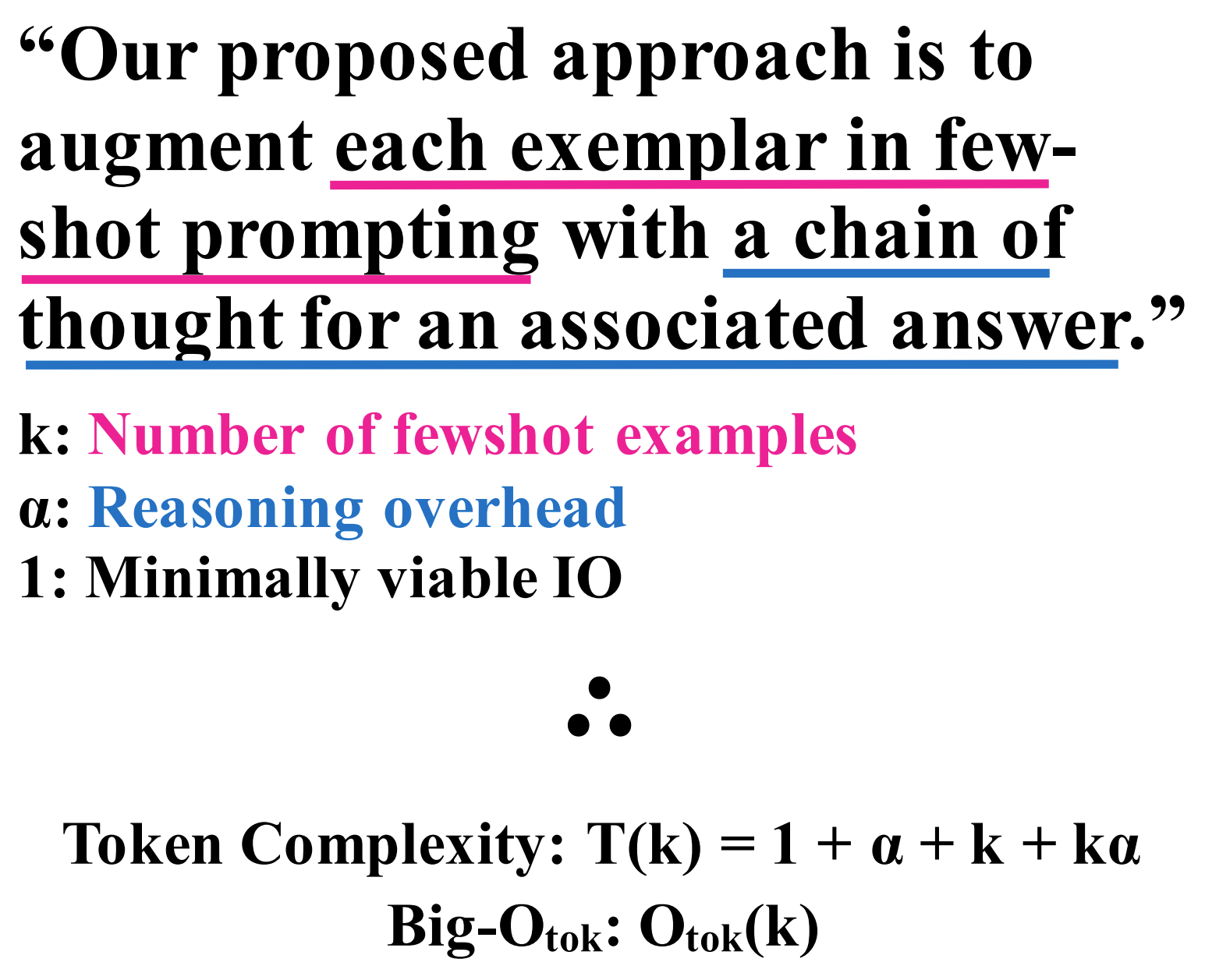}}
    \hspace{1cm}
    \subfloat[CoT-SC]{\label{fig:theor-deriv-f}\includegraphics[width=60mm]{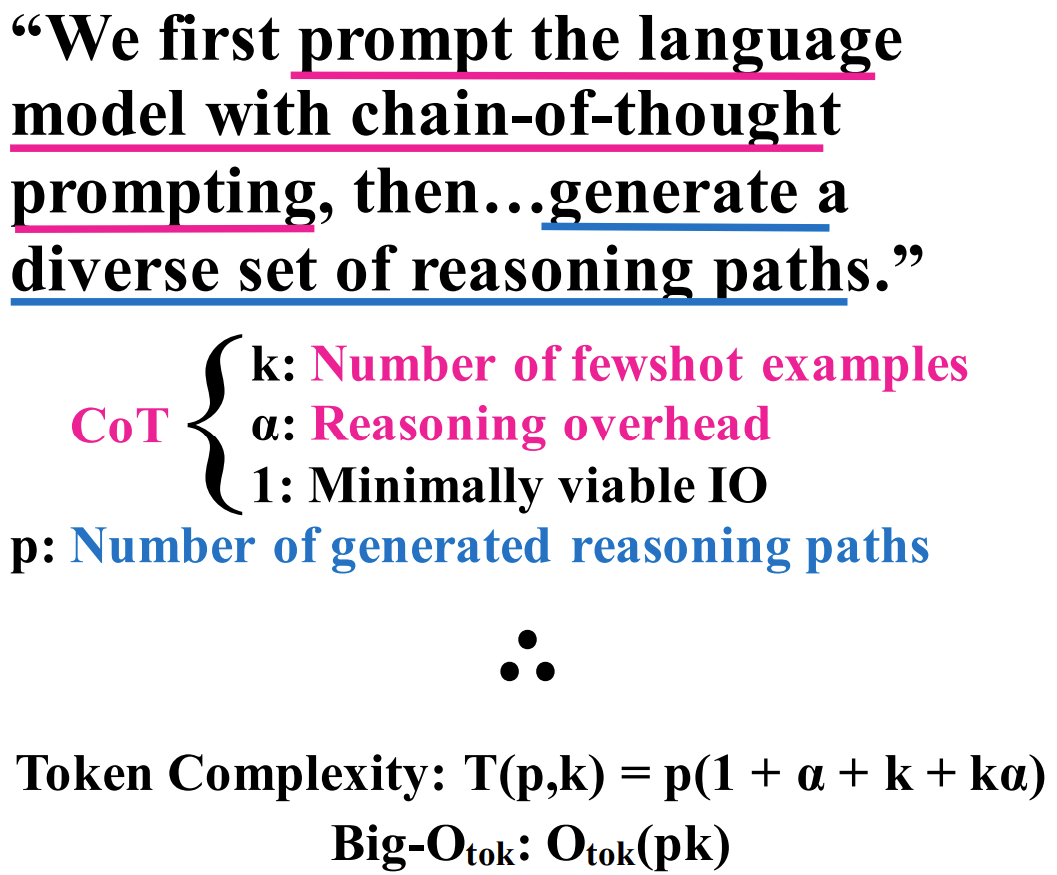}}
\caption{Sample derivations of \NAME. The textual descriptions in each figure are drawn from the following sources: (c) \cite{kojima2022largezeroshotcot}; (d) \cite{FewShotLearners}; (e) \cite{wei2022chain}; (f) \cite{wang2023selfconsistency}. Note that for (d), the fewshot examples are equivalent to the MVIO and we make the assumption that the LLM follows that pattern.}
\label{fig:theor-derivs}
\end{figure*}

We include examples of \NAME derivations for each prompting strategy examined in the main text in Figure \ref{fig:theor-derivs}.

%%%%%%%%%%%%%%%%%%%%%%%%%%%%%%%%%%%%%%%%%%%%%%%%%%%%%%%%%%%%%

\section{Interpreting \METRICNAME} \label{app:interpret-tc}

Although we include a thorough example of using \METRICABBREV for prompting strategy analysis, we seek to provide an expanded discussion on its interpretation here. For brevity, in the main text we state that, generally, low \METRICABBREV can be thought of as more efficient and high \METRICABBREV as less efficient. In most instances, this will hold true; however, there are some plausible edge cases that deserve consideration.

\subsection{Average \METRICABBREV}

One such edge case is exploiting extremely low token usages to achieve low average \METRICABBREV. For example, consider a multiple-choice benchmark with four options: (A), (B), (C), and (D). Assuming uniform distribution across the possible answers, a prompting strategy that would yield 25\% accuracy (assuming the LLM could consistently produce the correct output) might be: "Output (B)." At approximately 4 combined input and outputs tokens, such a prompting strategy would achieve an average \METRICABBREV of 0.16 $\frac{t}{p}$, a value much lower than any of our observed values in Table \ref{tab:all-results}. This demonstrates the need to test prompting strategies for generalizability. 

\subsection{Marginal \METRICABBREV}

For marginal \METRICABBREV, there are additional edge cases to consider. The formula for calculating marginal \METRICABBREV, \[\frac{num\ tokens_2 - num\ tokens_1}{accuracy_2 - accuracy_1}\] where $num\ tokens_2 >= num\ tokens_1$, allows for negative values. Despite being "low," a negative marginal \METRICABBREV value indicates extreme inefficiency since the prompting strategy will have consumed more tokens to achieve lower accuracy.

In the unlikely event that $num\ tokens_2 == num\ tokens_1$, marginal TC will not provide useful information. The more efficient prompting strategy would, in that case, be the one that attained higher accuracy. Similarly, if $accuracy_2 == accuracy_1$, marginal TC cannot be calculated, but the prompting strategy that consumed fewer tokens can be considered more efficient.

%%%%%%%%%%%%%%%%%%%%%%%%%%%%%%%%%%%%%%%%%%%%%%%%%%%%%%%%%%%%%

\section{Empirical Evaluation Details}

\subsection{Detailed Results}

We provide detailed results from our experiments in Table \ref{tab:all-results}. Note that the results presented for each prompting strategy, model, and benchmark combination are from a single execution with the number of samples noted in Table \ref{tab:bench-config}.

We removed empty outputs and outputs more than four standard deviations from the mean (e.g., instances where the LLM generated a looping output) from our token usage statistics. Such erroneous outputs were surprisingly common for Llama 3.1 8B Instruct. Details of the number of outputs removed from consideration are detailed in Table \ref{tab:removed}.

\begin{table*}
  \centering
  \small
  \begin{tabular}{c|c|c|c|c|c|c|c}
    \hline
     \multirow{2}{*}{\textbf{Model}} & \multirow{2}{*}{\textbf{Benchmark}} & \multicolumn{6}{c}{\textbf{Percentage of Total IO Pairs Excluded}} \\\cline{3-8}
     & & \rotatebox{90}{Vanilla IO} & \rotatebox{90}{Vanilla Fewshot} & \rotatebox{90}{Zeroshot CoT} & \rotatebox{90}{Fewshot CoT} & \rotatebox{90}{CoT-SC\textsubscript{5}} & \rotatebox{90}{CoT-SC\textsubscript{10}} \\\mytoprule
     \multirow{3}{*}{Llama 3.1 8B Instruct} & BBH & 2.53 & 8.59 & 2.90 & 5.04 & 4.09 & 4.00 \\\cline{2-8}
      & GSM8K & 0.68 & 1.14 & 0.76 & 0.91 & 1.27 & 1.28 \\\cline{2-8}
       & MMLU & 1.18 & 1.76 & 1.89 & 2.35 & 1.89 & 1.93 \\\hline
     \multirow{3}{*}{Qwen 2.5 14B Instruct} & BBH & 0.29 & 0.12 & 0.23 & 0.49 & 0.24 & 0.19 \\\cline{2-8}
      & GSM8K & 0.30 & 0.45 & 0.23 & 0.23 & 0.36 & 0.09 \\\cline{2-8}
       & MMLU & 0.07 & 0.65 & 0.20 & 0.46 & 0.20 & 0.38 \\\hline
    \multirow{3}{*}{Qwen 2.5 32B Instruct} & BBH & 0.22 & 0.09 & 0.26 & 0.29 & 0.16 & 0.10 \\\cline{2-8}
       & GSM8K & 0.00 & 0.00 & 0.08 & 0.38 & 0.18 & 0.13 \\\cline{2-8}
       & MMLU & 0.20 & 0.46 & 0.07 & 0.07 & 0.10 & 0.11 \\\mytoprule
  \end{tabular}
  \caption{The percentage of total IO pairs removed from token usage statistics. Pairs were removed if the output was empty or the length of the output was more than 4 standard deviations from the mean.}
  \label{tab:removed}
\end{table*}

\begin{table*}
  \centering
  \small
  \begin{tabular}{c|c|c|c|c|c|c|c|c}
    \hline
     & & \textbf{Strategy} & \textbf{Avg. Tokens\textsubscript{In}} & \textbf{Avg. Tokens\textsubscript{Out}} & \textbf{Avg. Tokens\textsubscript{Total}} & \textbf{Acc.$^*$} & \textbf{Std. Error} & \textbf{Average TC} \\\mytoprule
    \multirow{18}{*}{\rotatebox{90}{Llama 3.1 8B Instruct}} & \multirow{6}{*}{\rotatebox{90}{BBH}} & Vanilla IO & 172 & 221 & 393 & 51.1 & 0.56 & 7.70 \\\cline{3-9}
     &  & Vanilla 3-shot & 420 & 226 & 646 & 35.4 & 0.54 & 18.28 \\\cline{3-9}
     &  & Zeroshot CoT & 178 & 351 & 530 & 63.2 & 0.55 & 8.38 \\\cline{3-9}
     &  & 3-shot CoT & 876 & 335 & 1212 & 70.5 & 0.51 & 17.20 \\\cline{3-9}
     &  & CoT-SC\textsubscript{5} & 4378 & 1089 & 5468 & 72.7 & 0.50 & 75.19 \\\cline{3-9}
     &  & CoT-SC\textsubscript{10} & 8758 & 2177 & 10935 & 73.1 & 0.50 & 149.58 \\\cline{2-9}
     & \multirow{6}{*}{\rotatebox{90}{GSM8K}} & Vanilla IO & 122 & 162 & 284 & 77.3 & 1.15 & 3.68 \\\cline{3-9}
     &  & Vanilla 8-shot & 639 & 147 & 787 & 79.3 & 1.12 & 9.93 \\\cline{3-9}
     &  & Zeroshot CoT & 126 & 203 & 330 & 75.5 & 1.18 & 4.37 \\\cline{3-9}
     &  & 8-shot CoT & 654 & 145 & 800 & 83.4 & 1.02 & 9.60 \\\cline{3-9}
     &  & CoT-SC\textsubscript{5} & 3275 & 751 & 4026 & 85.7 & 0.96 & 46.96 \\\cline{3-9}
     &  & CoT-SC\textsubscript{10} & 6550 & 1499 & 8050 & 89.0 & 0.86 & 90.44 \\\cline{2-9}
     & \multirow{6}{*}{\rotatebox{90}{MMLU}} & Vanilla IO & 201 & 201 & 402 & 51.1 & 1.24 & 7.88 \\\cline{3-9}
     &  & Vanilla 4-shot & 711 & 208 & 920 & 58.8 & 1.19 & 15.65 \\\cline{3-9}
     &  & Zeroshot CoT & 207 & 342 & 550 & 69.4 & 1.13 & 7.93 \\\cline{3-9}
     &  & 4-shot CoT & 1008 & 182 & 1191 & 65.7 & 1.16 & 18.13 \\\cline{3-9}
     &  & CoT-SC\textsubscript{5} & 5037 & 955 & 5993 & 66.9 & 1.16 & 89.53 \\\cline{3-9}
     &  & CoT-SC\textsubscript{10} & 10076 & 1929 & 12006 & 69.5 & 1.12 & 172.76 \\\mytoprule
    \multirow{18}{*}{\rotatebox{90}{Qwen 2.5 14B Instruct}} & \multirow{6}{*}{\rotatebox{90}{BBH}} & Vanilla IO & 134 & 197 & 332 & 63.5 & 0.51 & 5.24 \\\cline{3-9}
     &  & Vanilla 3-shot & 379 & 143 & 523 & 62.0 & 0.55 & 8.44 \\\cline{3-9}
     &  & Zeroshot CoT & 140 & 254 & 395 & 79.7 & 0.42 & 4.96 \\\cline{3-9}
     &  & 3-shot CoT & 830 & 195 & 1026 & 81.2 & 0.43 & 12.64 \\\cline{3-9}
     &  & CoT-SC\textsubscript{5} & 4155 & 1011 & 5166 & 82.8 & 0.41 & 62.37 \\\cline{3-9}
     &  & CoT-SC\textsubscript{10} & 8310 & 2028 & 10339 & 83.7 & 0.40 & 123.52 \\\cline{2-9}
     & \multirow{6}{*}{\rotatebox{90}{GSM8K}} & Vanilla IO & 101 & 180 & 281 & 81.2 & 1.08 & 3.47 \\\cline{3-9}
     &  & Vanilla 8-shot & 618 & 150 & 768 & 83.5 & 1.02 & 9.21 \\\cline{3-9}
     &  & Zeroshot CoT & 104 & 194 & 299 & 81.9 & 1.06 & 3.65 \\\cline{3-9}
     &  & 8-shot CoT & 633 & 125 & 759 & 87.6 & 0.91 & 8.67 \\\cline{3-9}
     &  & CoT-SC\textsubscript{5} & 3168 & 602 & 3770 & 88.7 & 0.87 & 42.51 \\\cline{3-9}
     &  & CoT-SC\textsubscript{10} & 6337 & 1221 & 7559 & 89.7 & 0.84 & 84.28 \\\cline{2-9}
     & \multirow{6}{*}{\rotatebox{90}{MMLU}} & Vanilla IO & 162 & 162 & 325 & 76.4 & 1.05 & 4.26 \\\cline{3-9}
     &  & Vanilla 4-shot & 673 & 130 & 804 & 78.4 & 1.03 & 10.25 \\\cline{3-9}
     &  & Zeroshot CoT & 169 & 347 & 516 & 81.4 & 0.97 & 6.35 \\\cline{3-9}
     &  & 4-shot CoT & 965 & 163 & 1129 & 81.8 & 0.96 & 13.81 \\\cline{3-9}
     &  & CoT-SC\textsubscript{5} & 4829 & 883 & 5713 & 82.5 & 0.95 & 69.26 \\\cline{3-9}
     &  & CoT-SC\textsubscript{10} & 9650 & 1743 & 11394 & 82.9 & 0.94 & 137.47 \\\mytoprule
    \multirow{18}{*}{\rotatebox{90}{Qwen 2.5 32B Instruct}} & \multirow{6}{*}{\rotatebox{90}{BBH}} & Vanilla IO & 134 & 201 & 336 & 63.4 & 0.50 & 5.30 \\\cline{3-9}
     &  & Vanilla 3-shot & 379 & 144 & 523 & 71.2 & 0.49 & 7.36 \\\cline{3-9}
     &  & Zeroshot CoT & 141 & 242 & 383 & 82.3 & 0.39 & 4.66 \\\cline{3-9}
     &  & 3-shot CoT & 830 & 182 & 1012 & 87.2 & 0.37 & 11.62 \\\cline{3-9}
     &  & CoT-SC\textsubscript{5} & 4153 & 938 & 5092 & 88.5 & 0.36 & 57.56 \\\cline{3-9}
     &  & CoT-SC\textsubscript{10} & 8308 & 1883 & 10191 & 88.8 & 0.35 & 114.75 \\\cline{2-9}
     & \multirow{6}{*}{\rotatebox{90}{GSM8K}} & Vanilla IO & 101 & 197 & 298 & 83.4 & 1.02 & 3.58 \\\cline{3-9}
     &  & Vanilla 8-shot & 618 & 192 & 810 & 85.2 & 0.98 & 9.51 \\\cline{3-9}
     &  & Zeroshot CoT & 104 & 199 & 304 & 84.1 & 1.01 & 3.62 \\\cline{3-9}
     &  & 8-shot CoT & 633 & 135 & 769 & 89.4 & 0.85 & 8.60 \\\cline{3-9}
     &  & CoT-SC\textsubscript{5} & 3168 & 687 & 3856 & 89.8 & 0.83 & 42.96 \\\cline{3-9}
     &  & CoT-SC\textsubscript{10} & 6337 & 1373 & 7711 & 89.8 & 0.83 & 85.90 \\\cline{2-9}
     & \multirow{6}{*}{\rotatebox{90}{MMLU}} & Vanilla IO & 162 & 249 & 412 & 65.3 & 1.18 & 6.32 \\\cline{3-9}
     &  & Vanilla 4-shot & 671 & 194 & 865 & 77.3 & 1.05 & 11.21 \\\cline{3-9}
     &  & Zeroshot CoT & 169 & 357 & 526 & 84.5 & 0.89 & 6.23 \\\cline{3-9}
     &  & 4-shot CoT & 966 & 186 & 1153 & 85.2 & 0.87 & 13.53 \\\cline{3-9}
     &  & CoT-SC\textsubscript{5} & 4827 & 1013 & 5840 & 85.4 & 0.87 & 68.37 \\\cline{3-9}
     &  & CoT-SC\textsubscript{10} & 9652 & 2030 & 11682 & 86.1 & 0.86 & 135.70 \\\mytoprule
  \end{tabular}
  \par\tiny{$^*$ Accuracy as a percentage.}
  \caption{Detailed results from the empirical evaluation described in Section \ref{sec:prac-eval}.}
  \label{tab:all-results}
\end{table*}

\subsection{Additional Observations} \label{app:obs}

While our experiments were simple, we made a number of interesting observations that may warrant further analysis in future work.

The initial motivation behind fewshot prompting strategies was to define a pattern that the LLM would then follow in its answer \cite{FewShotLearners, wei2022chain}. For Fewshot CoT, this pattern included a reasoning chain that led to the right answer \cite{wei2022chain}. Now that LLMs are more capable and often aligned with human preferences after training, their default response to a question is to explain their reasoning before providing a response. This results in high output token usage even for Vanilla IO and Zeroshot CoT. Interestingly, \textbf{the reasoning chains (averaged, per benchmark) that the LLMs produced for Zeroshot CoT were longer in every instance than the ones produced for Fewshot CoT}, in some cases more than twice as long. Nonetheless, Fewshot CoT yielded accuracy improvements for nearly every benchmark. This supports the idea that the in-context learning of correct reasoning chains does positively influence the correctness of the generated reasoning chain, even if the LLM's default output is constrained.

That trend does not apply to Vanilla IO and Vanilla Fewshot, however. While Vanilla Fewshot outperforms Vanilla IO in almost every benchmark, the output token usage is less in most instances. This is likely caused by the pattern that is matched during Vanilla Fewshot, where the example outputs are the minimum number of tokens to convey the answer (e.g., for multiple-choice: "(A)").

We also observed how the quality of the fewshot reasoning chains provided as a part of Fewshot CoT affected performance. While Fewshot CoT yielded modest accuracy gains over Zeroshot CoT for BBH (4.6\%) and GSM8K (6.3\%), it actually registered a 0.9\% accuracy \textit{loss} on MMLU. This prompted an investigation into the CoT fewshot exemplars included in LM Evaluation Harness \cite{eval-harness}. The reasoning chains were significantly shorter than for the other two benchmarks. Interestingly, recent work on using concise reasoning chains, such as Constrained-CoT \cite{nayab2024concisethoughtsimpactoutput} and Concise-CoT \cite{renze2024benefitsconcisechainthought}, has demonstrated performance improvements from shorter chains of thought. An potentially insightful future work could explore how the form and content of intentionally concise chains of thought influence LLM performance to explain this discrepancy.

\subsection{Cost Estimates for Commercial Models} \label{app:commercial-cost}

\begin{table*}
  \centering
  \small
  \begin{tabular}{c|c|c|c|c|c}
    \hline
    \multirow{2}{*}{\textbf{Model}} & \multirow{2}{*}{\textbf{Price\textsubscript{input}$^*$}} & \multirow{2}{*}{\textbf{Price\textsubscript{output}$^*$}} & \multicolumn{3}{c}{\textbf{Cost$^\dagger$ (US\$)}} \\\cline{4-6}
     &  & & BBH & GSM8K & MMLU \\\mytoprule
    GPT-4o & 2.5 & 10.0 & 488.25 & 72.53 & 121.57 \\
    GPT-4o-mini & 0.15 & 0.6 & 29.29 & 4.35 & 7.29 \\
    Claude 3.5 Sonnet & 3.0 & 15.0 & 662.89 & 97.81 & 163.16 \\
    Claude 3.5 Haiku & 0.8 & 4.0 & 176.77 & 26.08 & 43.51 \\\mytoprule
  \end{tabular}
  \par\tiny{$^*$ Prices are in $\frac{\text{US\$}}{\text{1M Tokens}}$.}
  \par\tiny{$^\dagger$ Cost to run all prompting strategies on the given benchmark.}
  \caption{Cost estimates for recreating the empirical evaluation with common commercial models.}
  \label{tab:cost-estimate}
\end{table*}

The cost estimates found in Table \ref{tab:cost-estimate} are derived from the pricing pages for Anthropic\footnote{\url{https://anthropic.com/pricing\#anthropic-api}} and OpenAI\footnote{\url{https://openai.com/api/pricing/}}, accessed on January 31, 2025. We do not include the effects of prompt caching.

\subsection{Reproducibility} \label{app:repro}

All results\footnote{Provided under the CC-BY-4.0 license.}, as produced by LM Evaluation Harness, (e.g., LLM inputs and outputs, hyperparameters, runtimes, model configurations, etc.) are found at the following URL: \url{https://drive.google.com/file/d/1RsrzYUlrSuYzj43LRk33X2-Nru-7nG6y/view?usp=sharing}.

All code used to run the evaluations for this paper is found at the following GitHub repository: \url{https://github.com/Sypherd/lm-evaluation-harness/tree/reproduce-paper-results}. Although we used LM Evaluation Harness, we link our fork\footnote{Under the same MIT license as LM Evaluation Harness.} as significant bug fixes had to be made to get the framework to function as expected. Despite the bugs we encountered, we encourage others to support this open-source project that promotes reproducible results for LLM projects.

\subsection{Configurations} \label{app:configs}

\subsubsection{Models} \label{app:model-config}

We include details of model configurations in Table \ref{tab:model-config}. All models were sourced from Hugging Face\footnote{\url{https://huggingface.co/models}}. Where required, the authors complied with the necessary terms and conditions for gated models.

\begin{table*}
  \centering
  \small
  \begin{tabular}{c|c|c|c}
    \hline
    \textbf{Model$^*$} & \textbf{Max Context Length} & \textbf{Max Generations Tokens} & \textbf{Temperature} \\\mytoprule
    meta-llama/Llama-3.1-8B-Instruct & 128000 & 16384 & 0.0 (0.5 for CoT-SC)\\
    Qwen/Qwen2.5-14B-Instruct & 128000 & 8192 & 0.0 (0.5 for CoT-SC)\\
    Qwen/Qwen2.5-32B-Instruct & 128000 & 8192 & 0.0 (0.5 for CoT-SC)\\\mytoprule
  \end{tabular}
  \par\tiny{$^*$ Models were sourced from \url{https://huggingface.co/models}.}
  \caption{Model configurations used for the empirical evaluation.}
  \label{tab:model-config}
\end{table*}

\subsubsection{Benchmarks} \label{app:bench-config}

We include the benchmark configurations used for our experiments in Table \ref{tab:bench-config}. The underlying datasets for BBH, GSM8K, and MMLU were sourced from Hugging Face\footnote{\url{https://huggingface.co/datasets}}. While the fewshot examples for BBH were drawn from the same split used for evaluation, care was taken to ensure that the fewshot examples did not overlap with the target question.

We note in the main text that we selected general-purpose LLM benchmarks for our experiments but recognize that GSM8K could be seen as targeted towards the math domain. While it is true that GSM8K is composed of questions that require basic math, the reasoning capabilities and basic world knowledge it probes are generally applicable. The focus of the benchmark is "properly interpreting a question and reasoning through the steps to solve it" \cite{gsm8k}, not to evaluate advanced math skills. We believe this justifies GSM8K's inclusion as a general-purpose benchmark.

We rely on the steps taken by the authors in creating BBH \cite{suzgun2022challengingbbh, srivastava2023beyond}, GSM8K \cite{gsm8k}, and MMLU \cite{hendryckstest2021mmlu} to ensure ethical dataset creation, including the mitigation of bias, offensive content, and personally identifying information. We refer readers to those papers for additional information on the breadth of representation in those benchmarks, such as demographic groups.

\begin{table}
  \centering
  \small
  \begin{tabular}{c|c|c|c}
    \hline
    \textbf{Benchmark} & \textbf{Split} & \textbf{Fewshot Split} & \textbf{\# Samples} \\\mytoprule
    BBH & test & test$^*$ & 6511 \\
    GSM8K & test & train & 1319 \\
    MMLU & validation & dev & 1531 \\\mytoprule
  \end{tabular}
  \tiny{$^*$ Sampled fewshot examples did not overlap with the current benchmark question.}
  \caption{Benchmark configurations used for the empirical evaluation.}
  \label{tab:bench-config}
\end{table}

\subsubsection{Prompts}

To limit the inadvertent effects of linguistic prompt engineering (see \citet{salinas-morstatter-2024-butterfly}), we avoided making adjustments to the prompts as they were constructed by default in LM Evaluation Harness, for consistency. As such, we did not include output formatting instructions and used the default prompt template for each model. We excluded output formatting instructions for two reasons: (1) to not interfere with the LLM's natural output for the benchmark question and (2) to allow the benefits of implicit formatting for fewshot strategies. We include an example of a raw text input and output pair for Qwen 2.5 14B on MMLU using Zeroshot CoT below:

\textbf{Input}
\begin{lstlisting}[breaklines=true,texcl]
<|im_start|>system\nThe following are multiple choice questions (with answers) about high school psychology.\n\n<|im_end|>\n<|im_start|>user\nQuestion: Why are psychoanalysts sometimes interested in talking with a patient about dreams?\n(A) Psychoanalysts discovered that neurotransmitter abnormalities sometimes cause dream disturbances.; (B) Since all people are striving for self-actualization, psychoanalysts look at the cognitive obstacles in dreams.; (C) Nonproductive counterconditioning behaviors are sometimes revealed in dreams.; (D) Some psychoanalysts believe that dream symbols represent unconscious conflicts.\nAnswer: Let's think step by step.<|im_end|>\n<|im_start|>assistant\n
\end{lstlisting}

\textbf{Output}
\begin{lstlisting}[breaklines=true]
To determine the correct answer, let's analyze each option in the context of psychoanalytic theory:\n\n(A) Psychoanalysts discovered that neurotransmitter abnormalities sometimes cause dream disturbances.\n- This statement is more aligned with biological psychology rather than psychoanalysis. Psychoanalysis does not focus on neurotransmitters or biological abnormalities as the primary cause of dream disturbances.\n\n(B) Since all people are striving for self-actualization, psychoanalysts look at the cognitive obstacles in dreams.\n- This statement aligns more with humanistic psychology, particularly the work of Carl Rogers and Abraham Maslow, who emphasized self-actualization. Psychoanalysis, as developed by Freud, does not focus on self-actualization or cognitive obstacles in dreams.\n\n(C) Nonproductive counterconditioning behaviors are sometimes revealed in dreams.\n- This statement is more aligned with behaviorism and the concept of counterconditioning, which is not a primary focus of psychoanalysis. Psychoanalysis does not typically discuss counterconditioning in the context of dreams.\n\n(D) Some psychoanalysts believe that dream symbols represent unconscious conflicts.\n- This statement aligns with the psychoanalytic perspective, particularly Freud's theory of dreams. Freud believed that dreams are a manifestation of unconscious desires and conflicts, often symbolized in dream content.\n\nGiven this analysis, the correct answer is:\n\n(D) Some psychoanalysts believe that dream symbols represent unconscious conflicts.
\end{lstlisting}

\subsubsection{Compute}

Experiments were run on A100 40GB, A100 80GB, and H100 GPUs, as availability permitted. Configurations were limited to 1 or 2 GPUs per experiment. The experiments were run in tensor parallel or data parallel configurations, depending on the size of the model and the number of GPUs used. The exact configurations per experiment, as well as exact wall times, are detailed in the reproducibility materials referenced in Section \ref{app:repro}. The approximate number of GPU hours across all GPUs was 95.

\subsection{Licensing}

We use a number of open-source artifacts in this work. We list the licenses for each in Table \ref{tab:license}. We verify that our usage was in accordance with the projects' licenses.

\begin{table*}
  \centering
  \small
  \begin{tabular}{c|c|c}
    \hline
    \textbf{Artifact} & \textbf{License} & \textbf{Notes} \\\mytoprule
    Llama 3.1 8B Instruct & Llama 3.1 Community License & Copyright © Meta Platforms, Inc. All Rights Reserved. \\
    Qwen 2.5 14B Instruct & Qwen LICENSE AGREEMENT & Version: September 19, 2024 \\
    Qwen 2.5 32B Instruct & Qwen LICENSE AGREEMENT & Version: September 19, 2024 \\
    LM Evaluation Harness & MIT License & Copyright (c) 2020 EleutherAI \\
    BBH & MIT License & Copyright (c) 2022 suzgunmirac \\
    GSM8K & MIT License & Copyright (c) 2021 OpenAI \\
    MMLU & MIT License & Copyright (c) 2020 Dan Hendrycks \\\mytoprule
  \end{tabular}
  \caption{Licenses for the artifacts used in this work.}
  \label{tab:license}
\end{table*}

\section{Additional Studies} \label{app:abl}

\subsection{Ablation Study on the Number of Fewshot Exemplars} \label{app:abl:fewshot}

\begin{figure*}
    \centering
    \begin{minipage}[b]{0.45\textwidth}
        \centering
        \includegraphics[width=\textwidth]{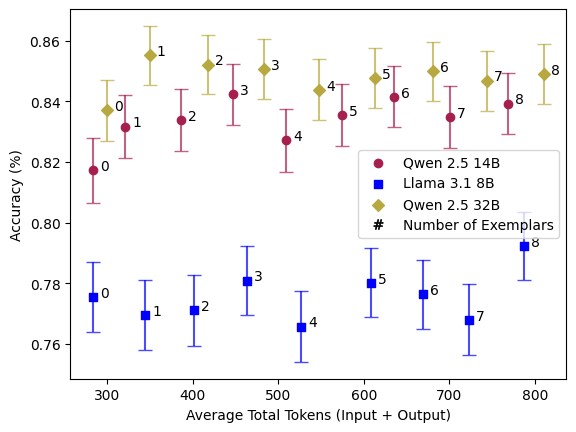}
        \small{(A) Vanilla Fewshot}
    \end{minipage}
    \begin{minipage}[b]{0.45\textwidth}
        \centering
        \includegraphics[width=\textwidth]{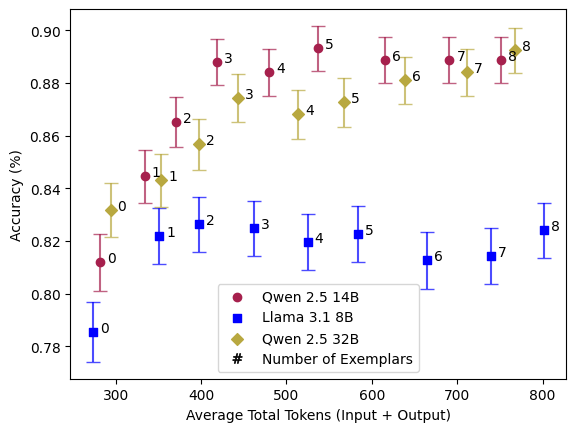}
        \small{(B) Fewshot CoT}
    \end{minipage}
    \caption{Accuracy and total token usage for the ablation study on the number of fewshot exemplars on the GSM8K benchmark. Standard error bars are included.}
    \label{fig:abl:fewshot}
\end{figure*}

We present the results from our ablation study on the number of fewshot examples in Figure \ref{fig:abl:fewshot}, with detailed results in Table \ref{tab:abl-fewshot}. For this experiment, we used the Vanilla Fewshot and Fewshot CoT prompting strategies with the number of exemplars ranging from 0 to 8. As can be seen in Figure \ref{fig:abl:fewshot}, the results are noisier but there is a clear trend of diminishing returns as the number of fewshot exemplars increases. To demonstrate this in a way that mitigates the noise, we examine the results piecewise, comparing the average marginal \METRICABBREV between 0 and 3 exemplars and 3 and 8 exemplars. Those values are found in Table \ref{tab:abl:fewshot-marginal}, for concision. The marginal TC between 0 and 3 fewshot exemplars is, for both Vanilla Fewshot and Fewshot CoT, over an order of magnitude less than between 3 and 8. This indicates a significant decrease in efficiency as token usage increases, which corroborates our findings in Section \ref{sec:results}.

\begin{table}
  \centering
  \small
  \begin{tabular}{c|c|c}
    \hline
    \multirow{2}{*}{\textbf{Fewshot Range}} & \multicolumn{2}{c}{\textbf{Marginal TC ($\frac{t}{p}$)}} \\\cline{2-3}
    & Vanilla Fewshot & Fewshot CoT \\\mytoprule
    0-3 & 117.2 & 30.5 \\
    3-8 & 1621.5 & 553.8 \\\mytoprule
  \end{tabular}
  \caption{Average marginal TCs ($\frac{\Delta tokens}{\Delta accuracy}$) calculated between 0 and 3 exemplars and 3 and 8 exemplars for the ablation study on the number of fewshot exemplars.}
  \label{tab:abl:fewshot-marginal}
\end{table}

\begin{table*}
  \centering
  \small
  \begin{tabular}{c|c|c|c|c|c|c|c|c|c|c}
    \hline
    \multirow{2}{*}{\textbf{Strategy}} & \multirow{2}{*}{\textbf{\# Exemplars}} & \multicolumn{3}{c|}{Llama 3.1 8B Instruct} & \multicolumn{3}{c|}{Qwen 2.5 14B Instruct} & \multicolumn{3}{c}{Qwen 2.5 32B Instruct} \\\cline{3-11}
    & & \textbf{Tokens} & \textbf{Acc.$^*$} & \textbf{SE$^\dagger$} & \textbf{Tokens} & \textbf{Acc.$^*$} & \textbf{SE$^\dagger$} & \textbf{Tokens} & \textbf{Acc.$^*$} & \textbf{SE$^\dagger$} \\\mytoprule
    \multirow{9}{*}{\rotatebox{90}{Vanilla Fewshot}} & 0 & 283.8 & 77.6 & 1.15 & 283.0 & 81.7 & 1.06 & 299.4 & 83.7 & 1.02 \\\cline{2-11}
    & 1 & 344.5 & 77.0 & 1.16 & 321.1 & 83.2 & 1.03 & 349.6 & 85.5 & 0.97 \\\cline{2-11}
    & 2 & 401.6 & 77.1 & 1.16 & 386.6 & 83.4 & 1.02 & 418.0 & 85.2 & 0.98 \\\cline{2-11}
    & 3 & 463.4 & 78.1 & 1.14 & 447.1 & 84.2 & 1.00 & 483.1 & 85.1 & 0.98 \\\cline{2-11}
    & 4 & 526.3 & 76.6 & 1.17 & 508.8 & 82.7 & 1.04 & 547.9 & 84.4 & 1.00 \\\cline{2-11}
    & 5 & 607.9 & 78.0 & 1.14 & 574.9 & 83.5 & 1.02 & 613.2 & 84.8 & 0.99 \\\cline{2-11}
    & 6 & 668.8 & 77.6 & 1.15 & 635.1 & 84.2 & 1.01 & 680.4 & 85.0 & 0.98 \\\cline{2-11}
    & 7 & 722.6 & 76.8 & 1.16 & 700.9 & 83.5 & 1.02 & 743.7 & 84.7 & 0.99 \\\cline{2-11}
    & 8 & 787.3 & 79.2 & 1.12 & 768.9 & 83.9 & 1.01 & 810.1 & 84.9 & 0.99 \\\mytoprule
    \multirow{9}{*}{\rotatebox{90}{Fewshot CoT}} & 0 & 273.2 & 78.5 & 1.13 & 281.7 & 81.2 & 1.08 & 294.2 & 83.2 & 1.03 \\\cline{2-11}
    & 1 & 351.3 & 82.2 & 1.05 & 334.1 & 84.5 & 1.00 & 353.7 & 84.3 & 1.00 \\\cline{2-11}
    & 2 & 398.1 & 82.6 & 1.04 & 370.5 & 86.5 & 0.94 & 397.7 & 85.7 & 0.97 \\\cline{2-11}
    & 3 & 462.6 & 82.5 & 1.05 & 418.9 & 88.8 & 0.87 & 443.1 & 87.4 & 0.91 \\\cline{2-11}
    & 4 & 524.8 & 82.0 & 1.06 & 479.6 & 88.4 & 0.88 & 514.2 & 86.8 & 0.93 \\\cline{2-11}
    & 5 & 584.3 & 82.3 & 1.05 & 537.4 & 89.3 & 0.85 & 567.8 & 87.3 & 0.92 \\\cline{2-11}
    & 6 & 664.6 & 81.3 & 1.07 & 615.6 & 88.9 & 0.87 & 638.9 & 88.1 & 0.89 \\\cline{2-11}
    & 7 & 739.2 & 81.4 & 1.07 & 691.0 & 88.9 & 0.87 & 711.0 & 88.4 & 0.88 \\\cline{2-11}
    & 8 & 801.3 & 82.4 & 1.05 & 751.9 & 88.9 & 0.87 & 768.3 & 89.2 & 0.85 \\\mytoprule
  \end{tabular}
  \par\tiny{$^*$ Accuracy as a percentage.}
  \par\tiny{$^\dagger$ Standard error.}
  \caption{Detailed results from the ablation study on the number of fewshot exemplars. Token counts represent the total token usage (input and output).}
  \label{tab:abl-fewshot}
\end{table*}

\subsection{Model Size} \label{app:abl:size}

\begin{figure*}
    \centering
    \begin{minipage}[b]{0.32\textwidth}
        \centering
        \includegraphics[width=\textwidth]{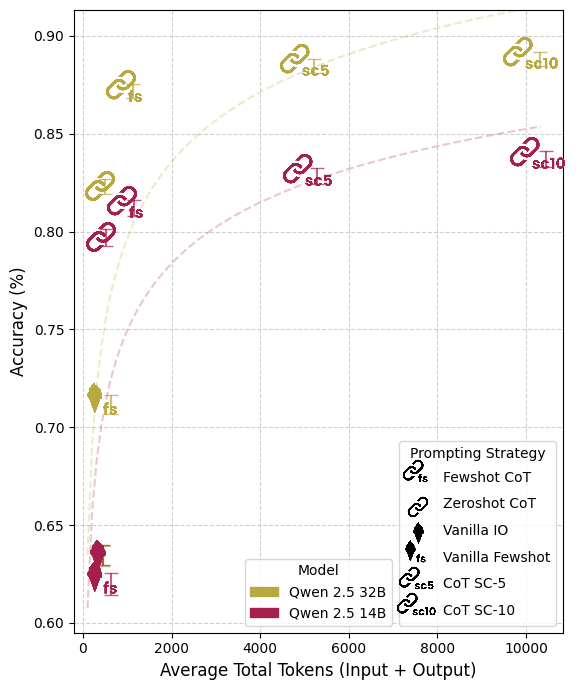}
        \small{(A) BBH}
    \end{minipage}
    \begin{minipage}[b]{0.32\textwidth}
        \centering
        \includegraphics[width=\textwidth]{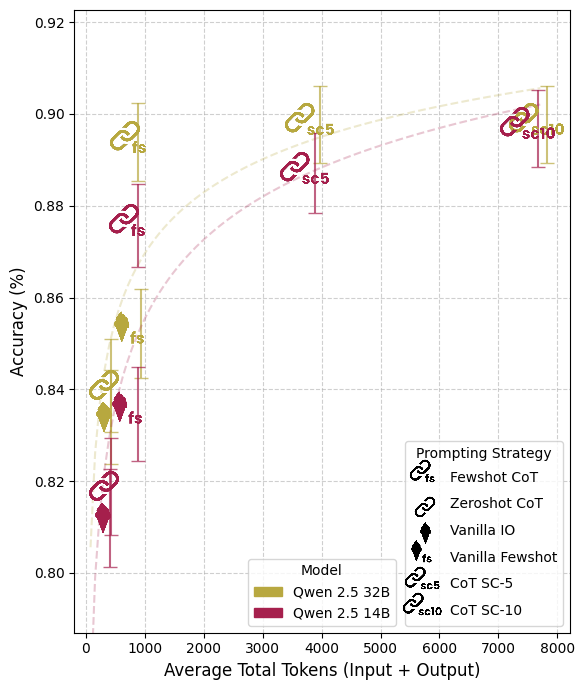}
        \small{(B) GSM8K}
    \end{minipage}
    \begin{minipage}[b]{0.32\textwidth}
        \centering
        \includegraphics[width=\textwidth]{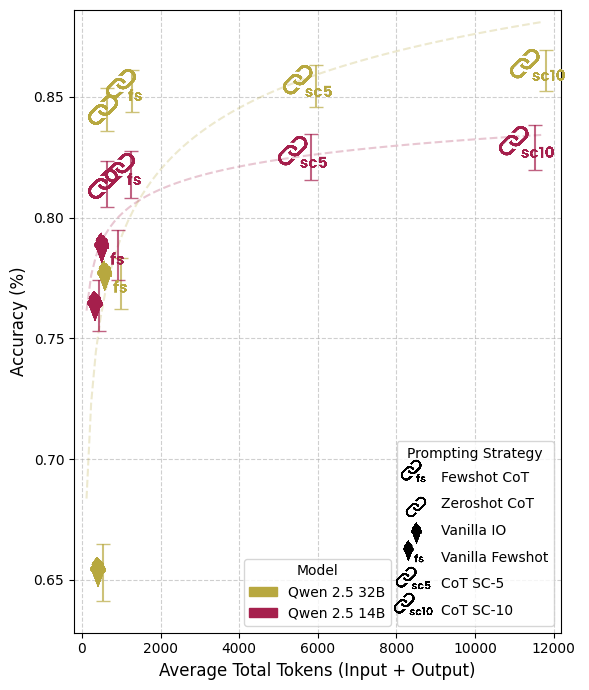}
        \small{(C) MMLU}
    \end{minipage}
    \caption{Accuracy and total token usage information for Qwen 2.5 14B and Qwen 2.5 32B from the empirical evaluation. The trend lines demonstrate the rapid growth of \METRICABBREV for these prompting strategies.}
    \label{fig:abl:model-size}
\end{figure*}

We included two models from the Qwen 2.5 family to facilitate a discussion on the impact of model size on our observed trends. Selecting two models from the same family ensures that potentially confounding variables, such as differences in training, data collection, and alignment, are presumably kept the same. We present the results from our experiments in Figure \ref{fig:abl:model-size}\footnote{These plots are identical to those from Figure \ref{fig:three_plots} but with Llama 3.1 8B excluded, for ease of comparison.}.

We observe that the trend towards diminishing accuracy returns for increased token usage is consistent between the 14B and 32B models. As expected, the 32B model generally outperforms the 14B model. However, we note some instances for prompting strategies that consume fewer tokens, such as Vanilla IO and Fewshot, where the 14B model outperforms the larger one. This suggests that larger models may be able to use additional tokens more effectively, as reflected in the consistently higher accuracy for Fewshot CoT and CoT-SC, but struggle to perform better than smaller models when fewer tokens are provided as context. This provides a promising route for future work.

\section{Use of AI}

There was limited use of AI in the research and writing of this work. For writing, ChatGPT\footnote{\url{https://chatgpt.com/}} was used to rephrase several sentences (<5) and to help debug LaTeX and Kubernetes errors. GitHub Copilot\footnote{\url{https://github.com/features/copilot}} was used to help generate some of the plots. Some grammatical suggestions from Writefull's Overleaf integration\footnote{\url{https://www.overleaf.com/learn/how-to/Writefull_integration}} were considered and included. All AI outputs were thoroughly reviewed by the authors prior to inclusion. 

%%%%%%%%%%%%%%%%%%%%%%%%%%%%%%%%%%%%%%%%%%%%%%%%%%%%%%%%%%%%%
%%%%%%%%%%%%%%%%%%%%%%%%%%%%%%%%%%%%%%%%%%%%%%%%%%%%%%%%%%%%%

\end{document}